\documentclass{article}
\usepackage{ijcai25}

\usepackage{times}
\usepackage{soul}
\usepackage{url}
\usepackage[hidelinks]{hyperref}
\usepackage[utf8]{inputenc}
\usepackage[small]{caption}
\usepackage{graphicx}
\usepackage{amsmath}
\usepackage{amsthm}
\usepackage{booktabs}
\usepackage{algorithm}
\usepackage{algorithmic}
\usepackage[switch]{lineno}

\usepackage{color}
\usepackage{xcolor}
\usepackage{subcaption}
\usepackage{newfloat}
\usepackage{listings}
\usepackage{amsmath}
\usepackage{makecell}
\usepackage{multirow}
\usepackage{amssymb}
\usepackage{bbding}

\usepackage{hyperref}
\hypersetup{hidelinks,
	colorlinks=true,
	allcolors=black,
	pdfstartview=Fit,
	breaklinks=true}

\title{Binary Event-Driven Spiking Transformer}

\author{
Honglin Cao$^{1}$
\and
Zijian Zhou$^{1}$\and
Wenjie Wei$^1$\and
Yu Liang$^1$\and
Ammar Belatreche$^2$\and
Dehao Zhang$^1$\and
Malu Zhang$^{1}$\footnotemark[1]\and
Yang Yang$^1$\And
Haizhou Li$^{3,4}$\\
\affiliations
$^1$University of Electronic Science and Technology of China\\
$^2$Northumbria University\\
$^3$The Chinese University of Hong Kong, Shenzhen\\
$^4$National University of Singapore\\
\emails
\{honglincao, zijianzhou\}@std.uestc.edu.cn,
maluzhang@uestc.edu.cn
}
\begin{document}

\maketitle

\renewcommand{\thefootnote}{\fnsymbol{footnote}}
\footnotetext[1]{Corresponding author.}

\begin{abstract}
    Transformer-based Spiking Neural Networks (SNNs) introduce a novel event-driven self-attention paradigm that combines the high performance of Transformers with the energy efficiency of SNNs. However, the larger model size and increased computational demands of the Transformer structure limit their practicality in resource-constrained scenarios. In this paper, we integrate binarization techniques into Transformer-based SNNs and propose the Binary Event-Driven Spiking Transformer, i.e. BESTformer. The proposed BESTformer can significantly reduce storage and computational demands by representing weights and attention maps with a mere 1-bit. However, BESTformer suffers from a severe performance drop from its full-precision counterpart due to the limited representation capability of binarization. To address this issue, we propose a Coupled Information Enhancement (CIE) method, which consists of a reversible framework and information enhancement distillation. By maximizing the mutual information between the binary model and its full-precision counterpart, the CIE method effectively mitigates the performance degradation of the BESTformer. Extensive experiments on static and neuromorphic datasets demonstrate that our method achieves superior performance to other binary SNNs, showcasing its potential as a compact yet high-performance model for resource-limited edge devices. The repository of this paper is available at \href{https://github.com/CaoHLin/BESTFormer}{https://github.com/CaoHLin/BESTFormer}.
\end{abstract}

\section{Introduction}
Spiking Neural Networks (SNNs) have attracted significant attention as third-generation artificial neural networks, known for their high biological plausibility and low power consumption \cite{maass1997networks}.
The spiking neuron utilizes binary spikes as the fundamental units for information transmission and works in a sparse spike-driven manner \cite{zhang2021rectified}.
This sparse synaptic transmission in SNNs simplifies multiply-accumulate (MAC) operations into accumulate (AC) operations, thereby significantly enhancing computational efficiency \cite{li2023brain,xu2024rsnn}. 
Furthermore, the energy-efficient feature of SNNs has driven the development of neuromorphic hardware, such as TrueNorth \cite{akopyan2015truenorth}, Loihi \cite{davies2018loihi}, and Tianjic \cite{pei2019towards}. 
However, despite the notable energy efficiency of SNNs, their performance in complex tasks still requires improvement.

\begin{figure}[t]
\centerline{\includegraphics[scale=0.3]{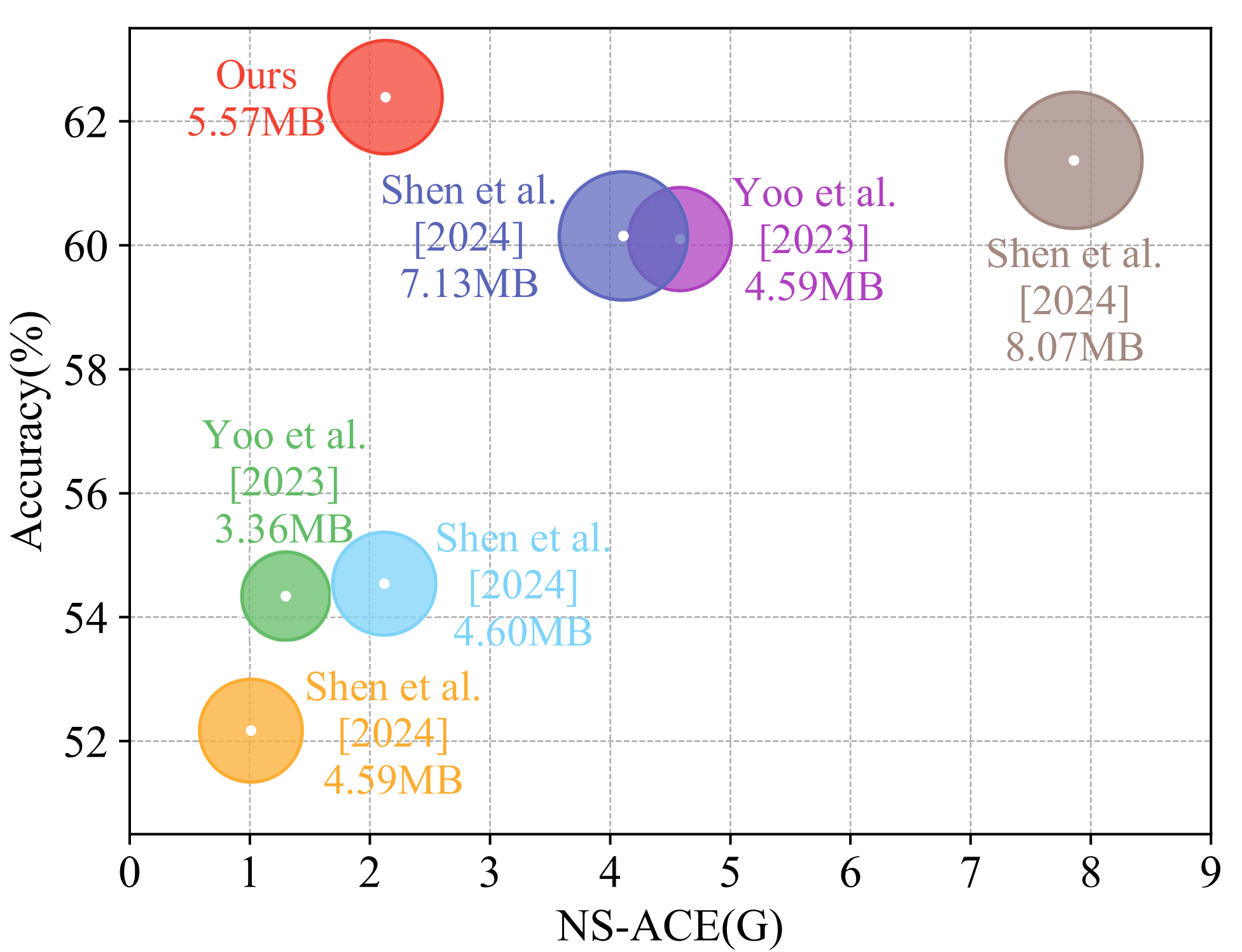}}
\caption{Accuracy vs. NS-ACE \& Model Size. Our method achieves superior computational and storage efficiency while outperforming other quantized SNNs on ImageNet. Neuromorphic Synaptic Arithmetic Computation Effort (NS-ACE) assesses SNN resource use in neuromorphic computing environments \protect\cite{shen2024conventional}.}
\label{fig:accuracy_sop}
\end{figure}

In recent years, there have been several studies integrating Transformers into SNNs, leading to a series of high-performance models, such as Spikformer \cite{zhou2023spikformer}, Spikingformer \cite{zhou2023spikingformer}, Spike-Driven Transformer v1 and v2 \cite{yao2024spikedriven,yao2024spike}, and SNN-ViT \cite{wang2025spiking}.
Compared to convolutional architectures in SNNs, these Transformer-based models have demonstrated significant performance improvements \cite{zhang2022spiking}.
However, their advancements typically rely on large model size, which is accompanied by substantial memory storage and computational overhead, limiting their deployment on resource-constrained edge devices.
Therefore, there is an urgent need for a compact yet high-performance Transformer-based SNN.

Quantization is a highly effective method for compressing large-scale models, which reduces model parameters from 32-bit to a low bit-width representation \cite{wei2025qp}. 
As an extreme form of quantization, binarization maximizes model size compression and accelerates computational speed by employing bitwise operations \cite{qin2020binary}.
Therefore, incorporating binarization with Transformer-based SNNs is promising for achieving an efficient and high-performance model.
It is worth noting that current research on binarization in SNN domains is primarily focused on convolutional structures, while Transformer-based structures remain unexplored \cite{yin2024mint,wei2024q,liang2025towards}.

In this paper, we explore the application of binarization technique in Transformer-based SNNs and propose a Binary Event-Driven Spiking Transformer (BESTformer) to minimize the model size and computational cost. 
Despite its high efficiency, the proposed BESTformer suffers from a significant performance drop due to limited information representation capability of binarization.
To address this issue, we propose the Coupled Information Enhancement (CIE) method to maximize the mutual information between the binary model and its full precision counterpart as much as possible.
By utilizing the CIE method in BESTformer, we improve its performance significantly while maintaining its efficiency advantage, as shown in Figure~\ref{fig:accuracy_sop}.
The main contributions of this paper are summarized as follows:
\begin{itemize}

\item We explore the combination of binarization with high-performance, low-power event-driven self-attention paradigm, proposing the Binary Event-Driven Spiking Transformer (BESTformer). The proposed BESTformer compresses both the weight parameters and attention map into mere 1-bit representations, aiming to reduce the model size and the excessive computational burden of Transformer-based SNNs.

\item We identify and analyse the performance degradation in BESTformer, which we attribute to the constrained information representation capability caused by binarization. Inspired by information theory, we propose the CIE method. This method utilizes a reversible framework and information enhancement distillation to maximize the mutual information between BESTformer and its full precision counterpart, leading to enhanced performance.

\item We conduct extensive experiments on static and neuromorphic datasets and demonstrate that the proposed BESTformer with the CIE method outperforms other binary SNNs. 
It's important to note that our method achieves a 7.85\% performance improvement on ImageNet-1k datasets compared to other models of similar scale at a time step of 1.
\end{itemize}
\section{Related Works}
\subsection{Transformer-based SNNs}
In recent years, there have been several studies integrating Transformer architectures into SNNs, leading to a series of high-performance SNN models. 
Spikeformer \cite{li2022spikeformer} is the first to integrate the Transformer architecture with SNNs, however, it retains numerous floating-point operations, making it unsuitable for neuromorphic computation. 
Spikformer \cite{zhou2023spikformer} introduces the Spiking Self Attention (SSA) mechanism, which enhances both energy efficiency and performance of Transformer-based SNNs.
Based on this, Spikingformer \cite{zhou2023spikingformer} modifies the residual connection within it to achieve a purely spike-driven Vision Transformer, further enhancing model's efficiency.
Spike-driven Transformer \cite{yao2024spike} proposes a spike-driven self-attention mechanism with linear complexity, significantly reducing energy consumption.
Then, to ensure versatility and high performance across various vision tasks, \cite{yao2024spikedriven} expand the original architecture into Meta-SpikeFormer, also known as Spike-driven Transformer v2.
Moreover, SNN-ViT \cite{wang2025spiking} achieves state-of-the-art performance in spiking vision tasks by introducing a saccadic self-attention mechanism specifically designed for spatio-temporal spike trains, maintaining linear computational complexity for edge applications.
Despite much progress, these models are limited by substantial memory and computational overheads, underscoring the need for further compression to reach their full potential.

\subsection{Quantization techniques in SNNs}
Various approaches have been proposed to quantize SNNs to low-bits.
\cite{deng2021comprehensive} employs spatiotemporal backpropagation (STBP) to directly train quantized SNNs and introduces the alternating direction method of multipliers (ADMM) to solve the performance degradation caused by quantization.
Then, to further enhance the performance, \cite{yoo2023cbp} uses constrained backpropagation (CBP) with the Lagrangian function as an objective function to quantize SNNs in training.
As an extreme form of quantization, binarization has also been widely studied.
\cite{qiao2021direct} presents a weight-binarized SNN to efficiently process event-based data, addressing the training demand of neuromorphic hardware for event data.
Moreover, \cite{10.3389/fnins.2023.1225871} proposes the accuracy loss estimator and binary weight optimization to achieve ultra-low latency adaptive local binary SNNs, which reduce memory storage by over 20\% while still maintaining high recognition accuracy. 
Recently, \cite{wei2024q} introduces a quantized SNN (Q-SNN) that reduces both weight and membrane potential representation, they also propose a weight-spike dual regulation (WS-DR) method to enhance the performance of Q-SNN.
Despite the effectiveness of these binarization methods in SNNs, they mainly focus on spiking convolutional architectures, while Transformer-based SNNs have not been explored.

\section{Method}
In this section, we first introduce the construction of BESTformer, including the weight binarization and the attention binarization. Subsequently, we analyze the challenge of limited information representation capability in the binary model. In order to address this challenge, we take inspiration from the information theory and propose the CIE method, which encompasses a reversible framework and information enhancement distillation.
\begin{table}[]
\tabcolsep=0.44cm
\def\arraystretch{1.25}
\centering
\begin{tabular}{c|ccc}
\hline
$\text{Attn}$        & $\operatorname{Boolean}$ & HR-LIF   &SR-LIF\\ \hline
{[}4,0,0,0{]} & {[}1,0,0,0{]}   & {[}1,0,0,0{]} & {[}1,1,0,0{]}      \\
{[}1,5,0,0{]} & {[}1,1,0,0{]}   & {[}1,1,0,0{]} & {[}1,1,1,0{]}     \\
{[}0,3,1,0{]} & {[}0,1,1,0{]}   & {[}0,1,1,0{]} &{[}0,1,1,0{]}      \\ \hline
\end{tabular}
\caption{A simple example of binarizing $\text{Attn}$ with $\operatorname{boolean}$ function, HR-LIF, and SR-LIF. The number of split patches $N$, time step $T$, threshold $V_{th}$, and time constant $\tau$ are set to 1, 4, 1, 0.5, respectively.}
\label{table:sr-lif}
\end{table}
\subsection{Binary Event-Driven Spiking Transformer}
\subsubsection{Weight binarization}
Existing Transformer-based SNNs typically utilize the Leaky Integrate-and-Fire (LIF) model with the hard reset mechanism (HR-LIF), its dynamics can be described as the following discrete form:
\begin{equation}
\label{lif-u}
\tilde{\text{U}}_l[t]=\tau \text{U}_l[t-1]+ \text{X}_l[t],
\end{equation}
where $\tau$ is time constant factor, $\text{U}_l[t]$ is the membrane potential of neurons in layer $l$ at time $t-1$, $\tilde{\text{U}}_l[t]$ is its intermediate representation, and $\text{X}_l[t]\in \mathbb{R}^{C \times H \times W}$ is the input current. $\text{X}_l[t]$ is integrated by presynaptic neurons, described as:
\begin{equation}
\text{X}_l[t]=\mathcal{BN}(\text{W}_{l}\text{S}_{l-1}[t]),
\end{equation}
where $\mathcal{BN}$ represents batch normalization, $\text{W}_{l}$ is the 32-bit weight matrix, and $\text{S}_{l-1}[t]$ is binary spike activities. Once membrane potential $\text{U}_l[t]$ reaches its firing threshold $V_{th}$, the neurons will generate a spike which can be described as:
\begin{equation}
    \text{S}_l[t]=\left\{\begin{array}{ll}1,&\text{if}~~ \tilde{\text{U}}_l[t]\geq V_{th},\\0,&\text{otherwise.}\end{array}\right.
\end{equation}
Neurons reset their membrane potential after emitting a spike. Typically, we set the reset potential to 0:
\begin{equation}
\text{U}_l[t]=(1-\text{S}_l[t]) \cdot \tilde{\text{U}}_l[t].
\end{equation}
To further reduce storage and computation demands, we quantize $\text{W}_{l}$ into 1-bit representation through the following formulas~\cite{qin2020forward}:
\begin{equation}
     \hat{\text{W}}_l = \text{W}_l - \bar{\text{W}}_l, \quad \hat{\text{W}}_l^{\text{std}} = \frac{\hat{\text{W}}_l}{\sigma(\hat{\text{W}}_l)},
\end{equation}
\begin{equation}
\label{bw}
      \text{B}_w = \begin{cases} +1, & \text{if } w \geq 0, \\ -1, & \text{otherwise,}
    \end{cases}
    ~w\in \hat{\text{W}}_l^{\text{std}},
\end{equation}
where $\bar{\text{W}}_l$ is the mean value of $\text{W}_l$, $\sigma(\hat{\text{W}}_l)$ is the standard deviation of $\hat{\text{W}}_l$. 
According to these two formulas, it can be seen that $\text{B}_w$ has two features: zero mean and normalization, where zero mean maximizes the information entropy of weight and normalization can accelerate the convergence process \cite{salimans2016weight,qin2020forward}. 
Therefore, Equation~\ref{lif-u} can be rewritten as:
\begin{equation}
\label{equation4}
\text{U}_l[t]=\tau \text{U}_l[t-1]+ \mathcal{BN}(\text{B}_{\text{W}_{l}}\otimes \text{S}_{l-1}[t]),
\end{equation}
where $\otimes$ is efficient bitwise operations, theoretically offering a $32\times$ memory saving and speedup compared with 32-bit operations~\cite{rastegari2016xnor}.
\begin{figure}[t]
\centering
    \includegraphics[width=0.47\textwidth]{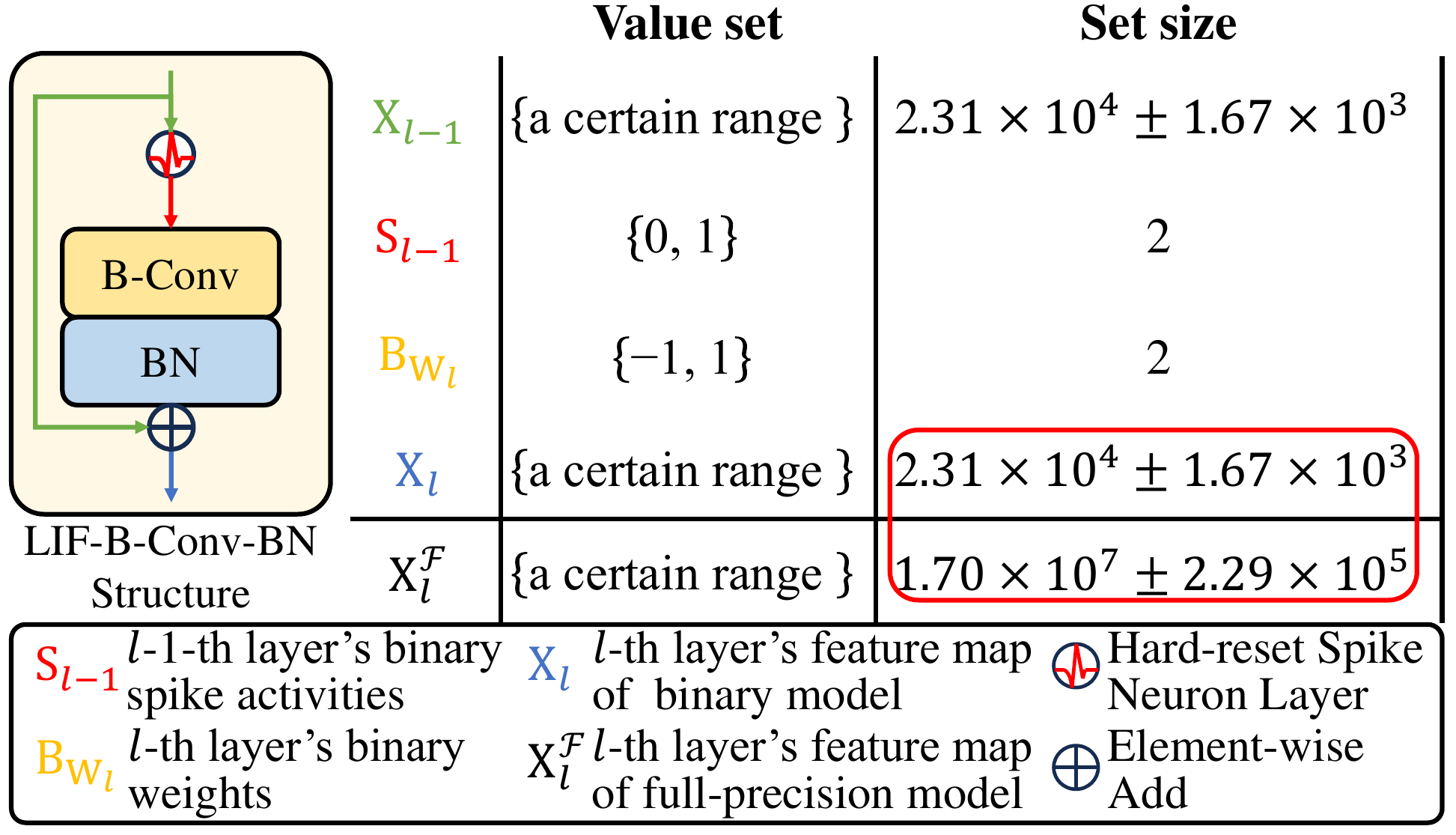}
    \caption{The `LIF-B-Conv-BN' structure of BESTformer and representation capability of variables in the structure. Value set indicates the collection of all values present in a variable. Set size indicates the size of a value set.}
\label{fig:representation capacity}
\end{figure}
\begin{figure*}
\centerline{\includegraphics[scale=0.50]{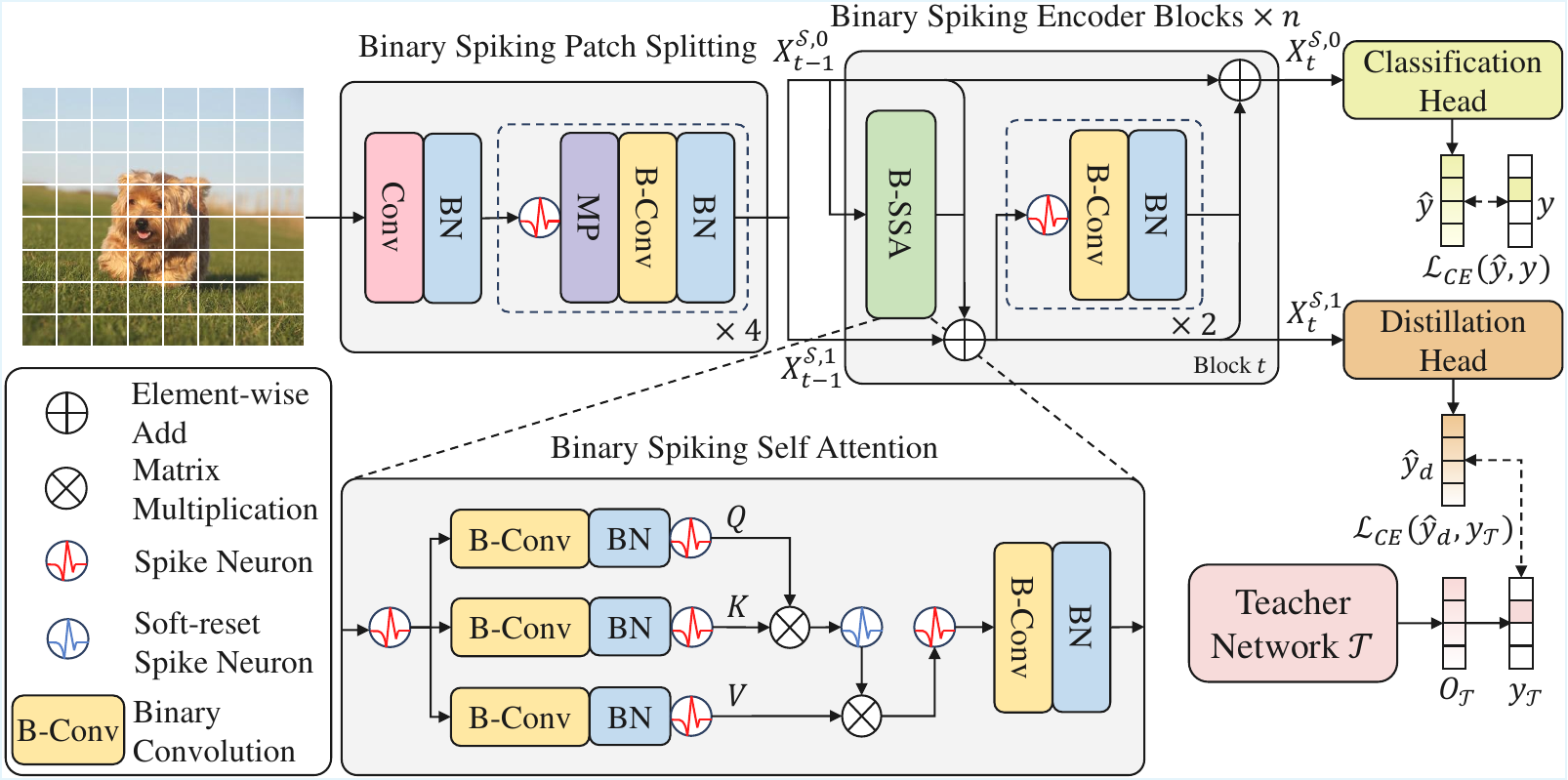}}
\caption{Overview of our BESTformer with the Coupled Information Enhancement method, which consists of a Binary Spiking Patch Splitting Module(BSPS), Reversible Binary Spiking Transformer Encoder Blocks, Classification and Distillation Heads.}
\label{fig:model}
\end{figure*}
\subsubsection{Attention binarization}
Aside from 1-bit weights and 1-bit spike activities, BESTformer also binarizes another crucial component in the Transformer-based SNNs, i.e., the attention map $\text{Attn}$.
Typically, $\text{Attn}$ is obtained through the matrix multiplication of two 1-bit spike vectors, Query $\text{Q}$ and Key $\text{K}$, yielding a non-negative integer result, described as:
\begin{equation}
\text{Attn} = \text{Q}\text{K}^\top,~~~\text{Attn}\in \mathbb{N}^{(T\times N\times N)},
\end{equation}
where $N$ is the number of split patches, $T$ is the time step of BESTformer.
To ensure full bitwise operations in BESTformer, we further binarize this attention map $\text{Attn}$.
However, directly binarizing it using the $\operatorname{boolean}$ or $\operatorname{sign}$ function will lead to limited information retention.
As the network depth increases, this limited information retention will result in severe performance degradation.
Fortunately, this issue can be alleviated by leveraging the additional temporal dimension of spiking neurons while maintaining the binary nature of BESTformer.
Given that each non-zero item in \text{Attn} is an integer at least 1, HR-LIF with a threshold of 1 will degenerate into the $\operatorname{boolean}$ function.
Therefore, in this paper, we use LIF with the soft reset mechanism (SR-LIF) to binarize $\text{Attn}$, mathematically defined as:
\begin{equation}
\text{B}_{\text{Attn}} =  \lambda \cdot \text{SR-LIF}(\text{Attn}),
\end{equation}
where $\lambda \in \mathbb{R}^{(T\times1\times1)}$ is the layer-wise learnable factors used to minimize binarization errors, which can be incorporated into the firing threshold during inference without requiring extra computations.
SR-LIF calculates the membrane potential after spike emission by subtracting the threshold, i.e., $\text{U}_l[t] = \tilde{\text{U}}_l[t]-\theta \text{S}_l[t]$.
As shown in Table~\ref{table:sr-lif}, when compared with $\operatorname{boolean}$ function or HR-LIF neurons, binarizing $\text{Attn}$ using SR-LIF can preserve more information.

\subsection{Challenge analysis}
BESTformer follows the event-driven `LIF-Conv-BN' design as commonly used in \cite{zhou2023spikingformer,yao2024spikedriven}, but replaces vanilla convolution with binary convolution. 
As shown in Figure~\ref{fig:representation capacity}, X is obtained by convolution and $\mathcal{BN}$ operations on binary $\text{B}_{\text{W}_l}$ and $\text{S}_{l-1}$. 

Previous research indicates that the poor performance of binarized neural networks is attributable to their low representational capability~\cite{liu2018bi,guo2022loss,guo2024ternary}.
In our research, we experimentally analyzed the average representation capability of the full-precision network and BESTformer on ImageNet-1k, respectively.
We use the value set size of feature maps as the measure of representation capability, which refers to the maximum number of distinct values in a feature map. The specific results are shown in Figure~\ref{fig:representation capacity}.

According to the huge difference in set size values of $\text{X}_{l}^{\mathcal{F}}$ and $\text{X}_{l}$,
we find that the representation capability of full-precision model is significantly greater than binary model.
This indicates that the information carried by BESTformer is severely constrained than its full-precision counterpart. 
This results in BESTformer losing a significant amount of useful information during the forward process, leading to unsatisfactory performance, especially in deep networks.

\subsection{Coupled information enhancement BESTformer}
Due to the constrained information representation capability of the binarized model, the ideal binary model should aim to retain the information representation of their full-precision counterparts as much as possible, thus the mutual information between the binarized and full-precision models' representations should be maximized~\cite{li2022q,xu2023q,xu2024reversing}.
Therefore, we propose a knowledge distillation framework, optimizing the mutual information $\mathcal{I}$ between the student model $\mathcal{S}$ and the full-precision teacher model $\mathcal{T}$, which can be formalized as:
\begin{equation} \max_{\theta^{\mathcal{S}}}\mathcal{I}(\text{X}_{n}^\mathcal{S};\text{X}_{m}^\mathcal{T}), \end{equation}
where $\theta^{\mathcal{S}}$ represents the parameters of the student model, and $\text{X}_{n}^\mathcal{S}$ and $\text{X}_{m}^\mathcal{T}$ correspond to the final ($n$-th and $m$-th) encoder blocks' outputs of the student and teacher models, respectively.
It's challenging to solve the maximization problem directly. Hence, we decompose this optimization objective into the difference between two entropy terms:
\begin{equation}
\mathcal{I}(\text{X}_{n}^\mathcal{S};\text{X}_{m}^\mathcal{T}) = \mathcal{H}(\text{X}_{n}^\mathcal{S}) - \mathcal{H}(\text{X}_{n}^\mathcal{S}|\text{X}_{m}^\mathcal{T}),
\end{equation}
where $\mathcal{H}(\mathbf{X})=-\int p(\boldsymbol{x})\log p(\boldsymbol{x})d\boldsymbol{x}$ and $p(\boldsymbol{x})$ denotes the probability density function of the random variable. We employed coupled information enhancement methods to optimize the objective: (1) maximizing $\mathcal{H}(\text{X}_{n}^\mathcal{S})$ to its upper bound by reversible framework \cite{gomez2017reversible}, and (2) minimizing $\mathcal{H}(\text{X}_{n}^\mathcal{S}|\text{X}_{m}^\mathcal{T})$ by information-enhanced distillation. The overall architecture diagram of applying the CIE method to BESTformer is shown in Figure~\ref{fig:model}.

\subsubsection{Reversible framework}
The information entropy of the model's feature maps exhibits a decreasing trend as the number of layers increases. This can be shown explicitly by the inequality in Proposition 1.
\paragraph{Proposition 1.}
\textit{In the context of deep neural networks, the information entropy of feature maps exhibits a non-increasing trend with respect to the depth of the network. That is, for} $\text{X}_l$ \textit{where} $l \in \{1, ..., n\}$, $\mathcal{H}(\text{X}_{l-1}) \geq \mathcal{H}(\text{X}_l)$ \textit{always holds true. Furthermore, we have}
$\mathcal{H}(\text{X}_0) \geq \mathcal{H}(\text{X}_1) \geq ... \geq \mathcal{H}(\text{X}_n)$.

\begin{figure}[t]
    \centering
    \begin{subfigure}[b]{0.36\textwidth}
        \centering
        \includegraphics[width=\textwidth]{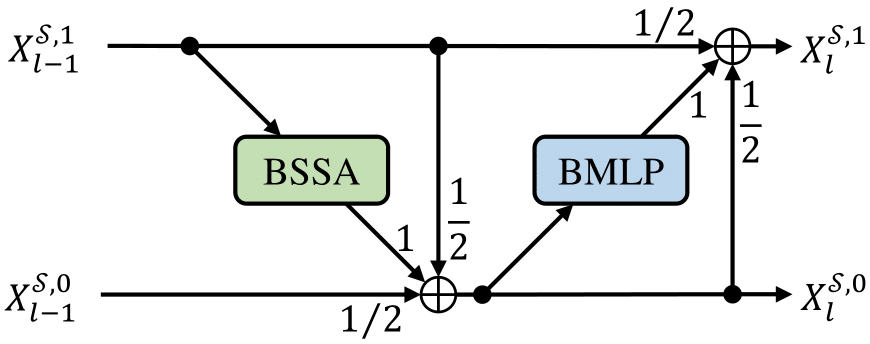}
        \caption{Forward}
        \label{fig:forward}
    \end{subfigure}
    \hspace{-2.9mm}
    \begin{subfigure}[b]{0.36\textwidth}
        \centering
        \includegraphics[width=\textwidth]{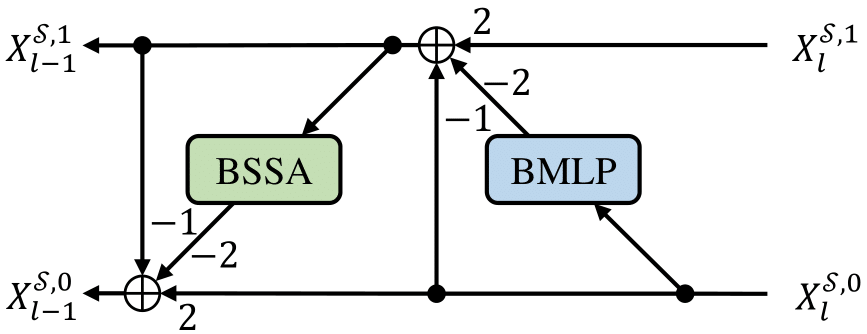}
        \caption{Inverse}
        \label{fig:inverse}
    \end{subfigure}
    \caption{Illustration of the forward and inverse process of the proposed reversible framework. The inverse process indicates that the inputs can be reconstructed from the outputs, i.e., this framework is reversible and no information is lost.}
    \label{fig:Reversible framework}
\end{figure}

Proposition 1 shows that information retained by the model tends to decrease as the neural network deepens, which is contrary to our objective of maximizing $\mathcal{H}(\text{X}_{n}^\mathcal{S})$. To address this issue, we utilize a reversible forward mapping in the encoder, as the reversible connection shown in Figure~\ref{fig:model}. We define forward mapping $\Phi_l(\text{X}_{l-1}^{\mathcal{S},0}, \text{X}_{l-1}^{\mathcal{S},1}) = (\text{X}_l^{\mathcal{S},0}, \text{X}_l^{\mathcal{S},1}) $ as:
\begin{equation}
\begin{aligned}
\text{X}_l^{\mathcal{S},0} &= \text{BSSA}(\text{X}_{l-1}^{\mathcal{S},1}) + \frac{1}{2}(\text{X}_{l-1}^{\mathcal{S},0} + \text{X}_{l-1}^{\mathcal{S},1}), \\
\text{X}_l^{\mathcal{S},1} &= \text{BMLP}(\text{X}_l^{\mathcal{S},0}) + \frac{1}{2}(\text{X}_{l-1}^{\mathcal{S},1} + \text{X}_l^{\mathcal{S},0}),
\end{aligned}
\end{equation}
where $\text{BSSA}$ denotes the Binary Spiking Self Attention and $\text{BMLP}$ represents the Binary MLP. As shown in Figure~\ref{fig:Reversible framework}, in the reversible encoder, the input can be accurately reconstructed from the output, ensuring that no information is lost in the process.
To verify this, we explicitly formulate the hidden inverse mapping that corresponds to the forward mapping as $\Phi_l^{-1}(\text{X}_l^{\mathcal{S},0}, \text{X}_l^{\mathcal{S},1}) = (\text{X}_{l-1}^{\mathcal{S},0}, \text{X}_{l-1}^{\mathcal{S},1})$ as:
\begin{equation}
\begin{aligned}
\text{X}_{l-1}^{\mathcal{S},1} &= 2(\text{X}_l^{\mathcal{S},1} - \text{BMLP}(\text{X}_l^{\mathcal{S},0})) - \text{X}_l^{\mathcal{S},0}, \\
\text{X}_{l-1}^{\mathcal{S},0} &= 2(\text{X}_l^{\mathcal{S},0} - \text{BSSA}(\text{X}_{l-1}^{\mathcal{S},1})) - \text{X}_{l-1}^{\mathcal{S},1}.
\end{aligned}
\end{equation}
Inverse mapping does not directly engage in network's operations, but it indirectly confirms that the information entropy within the reversible framework remains constant across all encoder blocks.
Therefore, $\mathcal{H}(\text{X}_n^\mathcal{S})$ reaches its upper bound $\mathcal{H}(\text{X}_0^\mathcal{S})$ as demonstrated by Proposition 2, which aligns with the objective of maximizing $\mathcal{H}(\text{X}_n^\mathcal{S})$. The detailed proofs of Proposition 1 and 2 are given in \textbf{Supplementary Materials}.
\paragraph{Proposition 2.}
\textit{In the context of reversible deep neural networks, the information entropy of the feature map remains invariant with respect to the depth of the network, i.e.} 
$\mathcal{H}(\text{X}_0) = \mathcal{H}(\text{X}_1) = ... = \mathcal{H}(\text{X}_n)$.

\subsubsection{Information enhanced distillation}
After modifying the Transformer Encoder Blocks of our binarized model to a reversible connection form, we maximize $\mathcal{H}(\text{X}_n^\mathcal{S})$ to its information entropy upper bound $\mathcal{H}(\text{X}_0^\mathcal{S})$. Maximizing $\mathcal{H}(\text{X}_0^\mathcal{S})$ can be achieved through standard network training and weight standardization. Consequently, the core optimization objective becomes:
\begin{equation} \min_{\theta^{\mathcal{S}}}\mathcal{H}(\text{X}_{n}^\mathcal{S}|\text{X}_{m}^\mathcal{T}). \end{equation}
Given the challenges associated with directly minimizing $\mathcal{H}(\text{X}_{n}^\mathcal{S}|\text{X}_{m}^\mathcal{T})$, we propose a knowledge distillation approach to implicitly achieve this minimization. Inspired by \cite{pmlr-v139-touvron21a}, we employ a dual-head architecture to fully leverage the information contained in the output features of the reversible network. Specifically, $\text{X}_{n}^{\mathcal{S},0}$ and $\text{X}_{n}^{\mathcal{S},1}$ are fed into the classification and distillation heads, respectively, generating the model's outputs $\hat{y}$ and $\hat{y_d}$. Let $O_{\mathcal{T}}$ denote the teacher model's output, and $y_{\mathcal{T}}=\arg\max{O_{\mathcal{T}}}$ represent the teacher model's `hard decision'. The global loss function incorporating distillation is formulated as:
\begin{equation} \mathcal{L}_{global} = (\mathcal{L}_{CE}(\hat{y}, y) + \mathcal{L}_{CE}(\hat{y_d}, y_{\mathcal{T}}))/2. \end{equation}
This dual-head design offers a significant advantage over traditional distillation methods by decoupling the gradient backpropagation for distillation and classification at the head level. This decoupling reduces mutual interference between the two gradients during backpropagation, facilitating more effective parameter updates.
It is worth noting that $\text{X}_{n}^{\mathcal{S},1}$, being the output of a deeper network compared to $\text{X}_{n}^{\mathcal{S},0}$, is more susceptible to overfitting when used for classification. Therefore, we strategically feed $\text{X}_{n}^{\mathcal{S},1}$ into the distillation head. This approach leverages the teacher model's output, which encapsulates `error information', thereby alleviating the overfitting problem and enhancing model's generalization capabilities.

\section{Experiments}

In this section, we first assess the classification performance of the proposed BESTformer with the CIE method on small-scale datasets, including CIFAR~\cite{krizhevsky2009learning}, CIFAR10-DVS~\cite{li2017cifar10}. Following this, we evaluate the method's performance on large-scale image dataset, ImageNet-1K~\cite{deng2009imagenet} to verify the scalability of our approach. Finally, we perform a series of ablation studies to validate the effectiveness of our method. The implementation details are provided in \textbf{Supplementary Materials}.

\begin{table*}[th]
\centering
\def\arraystretch{1.2}
\begin{tabular}{cccccccc}
\hline
Dataset                         & Method                                & Architecture        & \makecell[c]{Weight \\ Bits} & \makecell[c]{Time \\ Step} & \makecell[c]{Model Size\\(MB)($\downarrow$)} & Accuracy($\uparrow$) \\ \hline
\multirow{8}{*}{CIFAR-10}       & \multirow{2}{*}{\makecell[c]{\cite{yoo2023cbp}}}                         & VGG16 & 1                           & 32                          & 1.89          & 91.51\%  \\
& & VGG16 & 2                           & 32                          & 3.73          & 91.66\%  \\ \cline{2-7}
&   \cite{hu2024bitsnns}                      & ResNet18 & 1                           & 1                          & 1.42          & 93.74\%  \\
\cline{2-7}
                                & \multirow{2}{*}{\makecell[c]{\cite{shen2024conventional}}} & Spikformer-4-384    & 1                            & 4                          & 1.17           & 93.91\%  \\ 
                                &                                       & Spikformer-4-384    & 2                            & 2                          & 2.28           & 93.56\%  \\ \cline{2-7}
                                & \multirow{3}{*}{\makecell[c]{Ours}}                                  & Bestformer-2-384 & 1                            & 4                          & 0.73            & \textbf{94.98\%}  \\
                                &                                       & Bestformer-4-384 & 1                            & 2                          & 1.18           & \textbf{95.19\%}  \\
                                &                                       & Bestformer-4-384 & 1                            & 4                          & 1.18           & \textbf{95.73\%}  \\ \hline      
\multirow{8}{*}{CIFAR-100}      & \multirow{2}{*}{\makecell[c]{\cite{yoo2023cbp}}}                         & VGG16 & 1                           & 32                          & 2.08          & 66.53\%  \\
& & VGG16 & 2                           & 32                          & 3.90          & 66.46\%  \\ \cline{2-7}
&\cite{wei2024q} & ResNet19 & 1                           & 2                          & 1.56          & 78.77\%  \\ \cline{2-7}
                                & \multirow{2}{*}{\cite{shen2024conventional}} & Spikformer-4-384    & 1                            & 4                          & 1.24           & 74.13\%  \\
                                &                                       & Spikformer-4-384    & 2                            & 2                          & 2.34           & 75.91\%  \\ \cline{2-7}
                                & \multirow{3}{*}{\makecell[c]{Ours}}                                  & Bestformer-2-384 & 1                            & 4                          & 0.86           & \textbf{78.23\%}  \\
                                &                                       & Bestformer-4-384 & 1                            & 2                          & 1.31           & \textbf{79.23\%}  \\
                                &                                       & Bestformer-4-384 & 1                            & 4                          & 1.31           & \textbf{79.80\%}  \\ \hline  
\multirow{10}{*}{ImageNet} & \multirow{4}{*}{\makecell[c]{\cite{yoo2023cbp}}}                              & SEW-ResNet18        & 1                            & 4                          & 3.36                                  & 54.34\%  \\
& & SEW-ResNet18        & 2                            & 4                          & 4.88                                  & 58.04\%  \\
                                                       & & SEW-ResNet34        & 1                            & 4                          & 4.59                                  & 60.10\%  \\ 
& & SEW-ResNet34        & 2                            & 4                          & 7.13                                  & 62.98\%  \\ \cline{2-7}
& \multirow{4}{*}{\makecell[c]{\cite{shen2024conventional}}} & SEW-ResNet34        & 1                            & 1                          & 4.59                                  & 52.17\%  \\ 
                                                       & & SEW-ResNet34        & 2                            & 2                          & 7.13                                  & 60.15\%  \\ 
& & Spikformer-8-512    & 1                            & 1                          & 4.60                                  & 54.54\%  \\ 
                                                       & & Spikformer-8-512    & 2                            & 2                          & 8.07                                  & 61.37\%  \\ \cline{2-7}
& \multirow{2}{*}{Ours}                                & Bestformer-8-512 & 1                            & 1                          & 5.57                                 & \textbf{62.39\%}  \\
                                                       & & Bestformer-8-512 & 1                            & 4                          & 5.57                                 & \textbf{63.46\%}  \\ \hline
\multirow{5}{*}{CIFAR10-DVS}    & \multirow{2}{*}{\makecell[c]{\cite{yoo2023cbp}}}        & Wide-7B-Net & 1                           & 16                         & 0.17           & 74.70\%  \\
& & Wide-7B-Net & 2                           & 16                         & 0.32           & 75.30\%  \\ \cline{2-7}
                                & \makecell[c]{\cite{shen2024conventional}}                  & Spikformer-2-256    & 1                            & 16                         & 0.33           & 79.80\%  \\ \cline{2-7}
                                & \multirow{2}{*}{Ours}                 & Bestformer-2-256 & 1                            & 10                         & 0.34           & 78.70\%       \\
                                &                                       & Bestformer-2-256 & 1                            & 16                         & 0.34           & \textbf{80.80\%}       \\ \hline
\end{tabular}
\caption{Classification performance comparison on CIFAR-10, CIFAR-100, ImageNet and CIFAR10-DVS.}
\label{table:cifar}
\end{table*}

\subsection{Comparison with Related Work}
\subsubsection{Results on small-scale datasets classification}


We evaluate our BESTformer with the CIE method on small-scale datasets, including static datasets, CIFAR-10 and CIFAR-100, and neuromorphic datasets CIFAR10-DVS. 
The experimental results in Table~\ref{table:cifar} demonstrate the superior performance and efficiency of our proposed method across multiple benchmark datasets.

On the CIFAR-10 dataset, our 1-bit Bestformer-4-384 achieves a remarkable accuracy of 95.73\%, significantly outperforming previous state-of-the-art Transformer-based methods (93.91\%) and conventional Conv-based approaches (91.66\%). Similarly, on CIFAR-100, our 1-bit model attains 79.80\% accuracy, representing a substantial improvement of 3.89\% over the previous best result of 75.91\% achieved by 2-bit. Notably, this performance enhancement is achieved while maintaining exceptional model efficiency. Our 1-bit Bestformer-4-384 requires only 1.18MB of storage for CIFAR-10 and 1.31MB for CIFAR-100, representing more than a 93\% reduction in model size compared to full-precision counterparts. This demonstrates that our quantization approach effectively preserves model accuracy while significantly reducing memory requirements.

The robustness of our method is further validated on neuromorphic datasets. On CIFAR10-DVS, our 1-bit Bestformer-2-256 achieves 80.80\% accuracy, a state-of-the-art result. This consistent performance across both static and neuromorphic datasets underscores the versatility and effectiveness of our approach in different application scenarios.

\subsubsection{Results on ImageNet-1k classification}

On the challenging ImageNet dataset, Our model consistently shows superior performance compared to other quantization methods. According to Table~\ref{table:cifar}, our 1-bit Bestformer-8-512 achieves an impressive accuracy of 63.46\% with 4 time steps, outperforming other quantized methods such as CBP-QSNN (60.10\% with SEW-ResNet34) and Quantized Spikformer (61.37\% with 2-bit weights). 
Moreover, with just 1 time step, the model attains an accuracy of 62.39\%, improving performance by 7.85\% (54.54\% with 1-bit weight and 1 time step by Quantized Spikformer) while maintaining a similar model size and comparable computational complexity.

Compared to full-precision models, our method significantly reduces model size and computational efficiency. According to Table~\ref{table:energy}, our 1-bit Bestformer-8-512 model is 5.57MB, approximately 10 times smaller than its full-precision counterpart (59.36MB). In terms of computational efficiency, our 1-bit Bestformer-8-512 model with 4 time steps requires only 5.67G Neuromorphic Synaptic Arithmetic Computation Effort (NS-ACE~\cite{shen2024conventional}, more details are shown in \textbf{Supplementary Materials}), a 94.6\% reduction from the 104.32G NS-ACE of the full-precision model. This substantial decrease in NS-ACE indicates markedly improved energy efficiency of our model on neuromorphic hardware.


    

\subsection{Ablation study}

\begin{table}[t]
\centering
\begin{tabular}{c|ccc}

\hline
\multirow{2}{*}{Model} & \multirow{2}{*}{\begin{tabular}[c]{@{}c@{}}Model Size\\ (MB)($\downarrow$)\end{tabular}} & \multirow{2}{*}{\begin{tabular}[c]{@{}c@{}}SOPs\\ (G)($\downarrow$)\end{tabular}} & \multirow{2}{*}{\begin{tabular}[c]{@{}c@{}}NS-ACE\\ (G)($\downarrow$)\end{tabular}} \\
                                &                                                                                     &                                                                              &                                                                          \\ \hline
Full-precision$\dagger$                       & 59.36                                                                               & 6.52                                                                        & 104.32                                                                                                                                                  \\ \hline
Q-SEW-ResNet$\ddagger$                    & 7.13                                                                                & \textbf{2.06}                                                                         & 4.11                                                                                                                                                  \\
Q-Spikformer$\ddagger$                    & 8.07                                                                                & 3.93                                                                         & 7.86                                                                                                                                                  \\
Ours                            & \textbf{5.57}                                                                                & 2.13                                                                         & \textbf{2.13}                                                                                                                                                  \\ \hline

\end{tabular}
\caption{Comparison of resource consumption on ImageNet-1k. $\dagger$ The results for the full-precision model are taken from \protect\cite{zhou2023spikingformer}. $\ddagger$ The quantized results are taken from \protect\cite{shen2024conventional}.}
\label{table:energy}
\end{table}
\subsubsection{Impact of components of CIE on model accuracy}
To validate the effectiveness of our proposed components, we conduct comprehensive ablation studies on CIFAR-100. As shown in Table~\ref{table:ablation}, we progressively incorporate different components into our framework and observe their individual and combined effects. The baseline model achieves 77.77\% accuracy. Adding the reversible architecture (RF) brings a 0.50\% improvement, while incorporating information enhancement distillation (IED) alone leads to a more substantial 1.34\% gain. When combining both components, we explore two variants: using $X_l^{S,0}$ for distillation and $X_l^{S,1}$ for classification yields a 1.90\% improvement, while the reverse configuration 
achieves the best performance with a 2.03\% accuracy gain. These results demonstrate that both the reversible architecture and information enhancement distillation contribute positively to the model's performance, with their combination producing synergistic effects.

\begin{table}[b]
\centering
\begin{tabular}{c|cc}
\hline
Method        & \begin{tabular}[c]{@{}c@{}}Accuracy\\ (\%)\end{tabular} & \begin{tabular}[c]{@{}c@{}}Increment\\ (\%)\end{tabular} \\ \hline
baseline     & 77.77                                                   & -                                                        \\ \hline
w/ RF        & 78.27                                                   & + 0.50 
                                     \\
w/ IED        & 79.14                                                   & + 1.34                                                     \\
w/ RF \& IED$\dagger$ & 79.67                                                   & + 1.90                                                     \\
w/ RF \& IED$\ddagger$  & 79.80                                                   & + 2.03                                                     \\ \hline
\end{tabular}
\caption{Ablation study of CIE design on CIFAR-100. 
RF stands for reversible architecture, and IED stands for information enhancement distillation. Symbol $\dagger$ means that $X_l^{S,0}$ in Figure~\ref{fig:model} is used for distillation and $X_l^{S,1}$ is used for classification. Symbol $\ddagger$ means that $X_l^{S,0}$ is used for classification and $X_l^{S,1}$ is used for distillation.
}
\label{table:ablation}
\end{table}

\subsubsection{Impact of CIE on different architectures}
\begin{figure}[t]
\centering
    \begin{subfigure}[b]{0.243\textwidth}
        \centering
        \includegraphics[width=\textwidth]{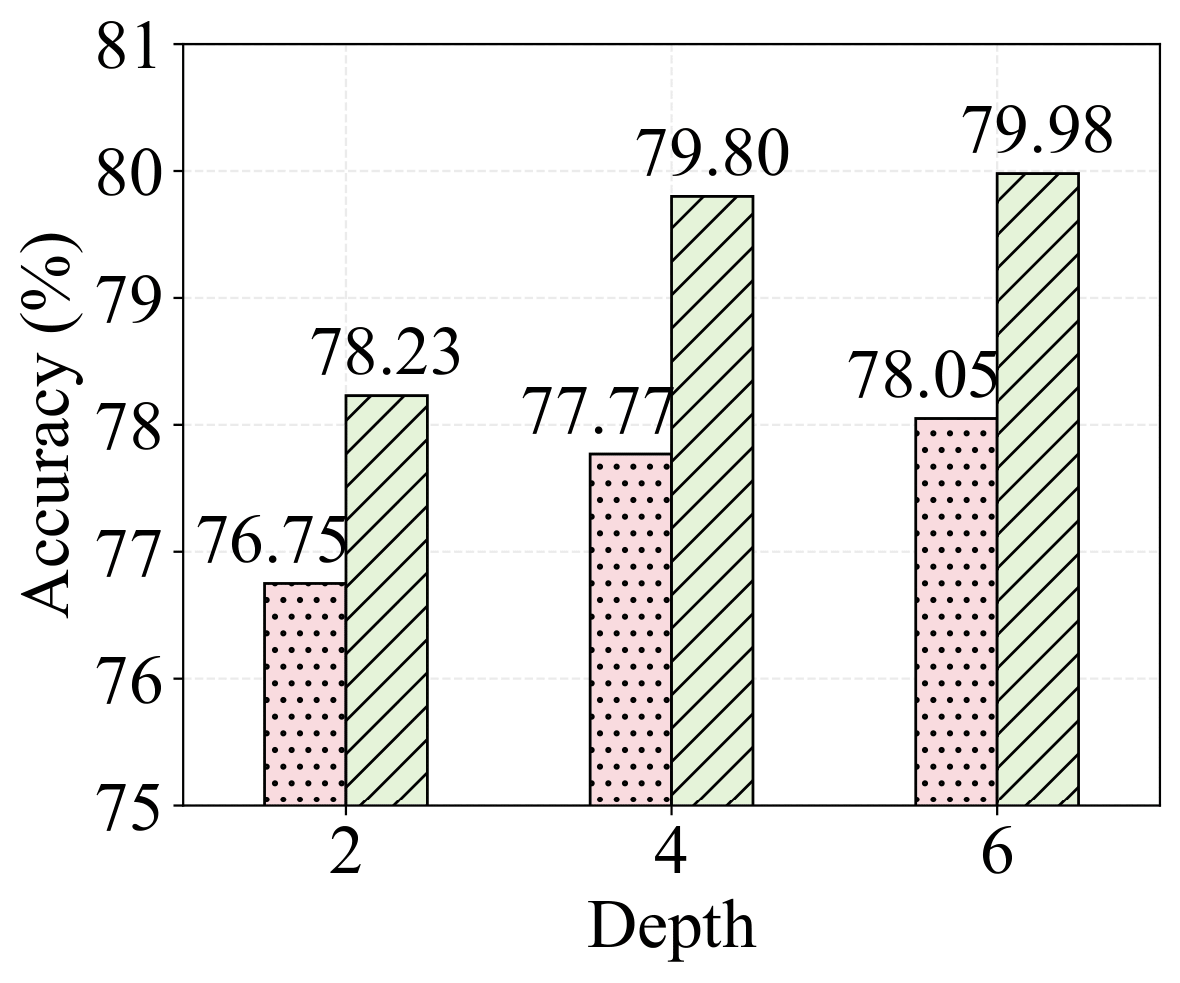}
        \caption{}
        \label{fig:ablation_accuracy_a}
    \end{subfigure}
    \hspace{-7pt}
    \begin{subfigure}[b]{0.243\textwidth}
        \centering
        \includegraphics[width=\textwidth]{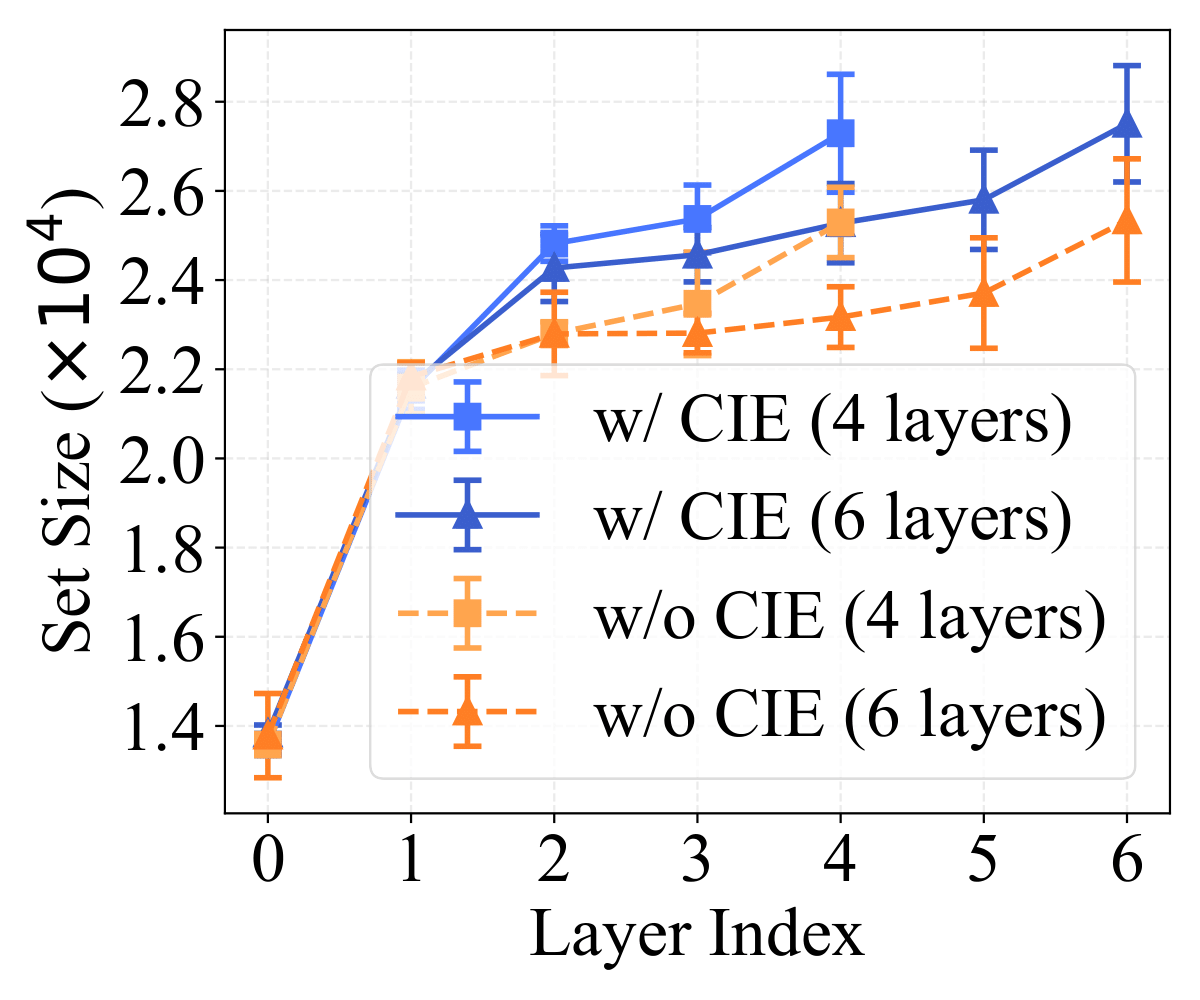}
        \caption{}
        \label{fig:ablation_accuracy_b}
    \end{subfigure}
    \caption{Ablation study for CIE method on CIFAR-100. (a) Impact of CIE on model accuracy across different architectures. (b) A comparative analysis of information representation capability: evaluating models with and without CIE method across variable architecture depths.}
\label{fig:ablation_accuracy}
\end{figure}

Our ablation experiments on CIFAR100 demonstrate the effectiveness of the CIE method across different Bestformer architectures. The results of these ablation experiments are presented in Figure~\ref{fig:ablation_accuracy_a}. The Bestformer-2-384 architecture achieved 76.75\% accuracy, which was significantly enhanced to 78.23\% with CIE integration. Similarly, Bestformer-4-384 exhibited improved performance from 77.77\% to 79.80\%, while Bestformer-6-384 achieved the highest overall accuracy of 79.98\% after applying CIE, compared to its baseline of 78.05\%. The consistent performance improvements across all architectures suggest that CIE effectively enhances feature representation regardless of model depth. Additionally, while deeper architectures demonstrated higher baseline performance, Bestformer-4-384 achieved the most substantial improvement (+2.03\%) with CIE integration, indicating an optimal balance between model capacity and enhancement effectiveness.

\subsubsection{Impact of CIE on information representation}

To further validate the effectiveness of CIE, we conducted an in-depth analysis of information representation capability across different encoder blocks on CIFAR-100 datasets as shown in Figure~\ref{fig:ablation_accuracy_b}.
We conducted multiple experiments to obtain the average values and corresponding error margins for all architectures to assess the overall trend. The results demonstrate that BESTformer with CIE method consistently maintains higher representation capability than original baseline (without CIE), especially in the later encoder blocks.
The improved information representation ability helped the model with 4 layer encoder blocks achieve a 2.03\% performance improvement.
The representation gap between them remains relatively stable across different architectures, underlining the consistent superiority of our approach.
These findings support the effectiveness of our CIE method, which contributes to improved performance in downstream tasks.

\section{Conclusion}
This work introduces a Binary Event-Driven Spiking Transformer that significantly reduces the storage and computational demands of Transformer-based Spiking Neural Networks. To address the constrained information representation capability caused by binarization, we propose the Coupled Information Enhancement (CIE) method, which combines a reversible framework and information enhancement distillation. Extensive experiments on both static and neuromorphic datasets demonstrate that BESTformer with CIE achieves superior performance compared to other binary SNNs while maintaining high efficiency. Our work provides a promising direction for developing compact yet high-performance models for resource-constrained edge devices. 

\section*{Acknowledgments}
This work is supported in part by the National Natural Science Foundation of China (No. U2333211, U20B2063 and 62220106008), in part by the Project of Sichuan Engineering Technology Research Center for Civil Aviation Flight Technology and Flight Safety (No. GY2024-27D), in part by the Open Research Fund of the State Key Laboratory of Brain-Machine Intelligence, Zhejiang University (Grant No.BMI2400020), in part by the Shenzhen Science and Technology Program (Shenzhen Key Laboratory, Grant No. ZDSYS20230626091302006), in part by the Shenzhen Science and Technology Research Fund (Fundamental Research Key Project, Grant No. JCYJ20220818103001002) and in part by the Program for Guangdong Introducing Innovative and Enterpreneurial Teams, Grant No. 2023ZT10X044.

\section*{Contribution Statement}
Honglin Cao and Zijian Zhou contribute equally to this work.

\bibliographystyle{named}
\bibliography{ijcai25}

\end{document}